\documentclass[conference]{IEEEtran}
\IEEEoverridecommandlockouts
\usepackage{cite}
\usepackage{amsmath,amssymb,amsfonts}
\usepackage{algorithmic}
\usepackage{graphicx}
\usepackage{textcomp}
\usepackage{xcolor}
\usepackage{float}
\usepackage{placeins}
\usepackage[colorlinks=true,
            linkcolor=black,
            citecolor=black,
            urlcolor=black]{hyperref}

\def\BibTeX{{\rm B\kern-.05em{\sc i\kern-.025em b}\kern-.08em
    T\kern-.1667em\lower.7ex\hbox{E}\kern-.125emX}}

\DeclareRobustCommand{\orcidicon}{%
  \raisebox{-0.25ex}{\includegraphics[width=0.9em]{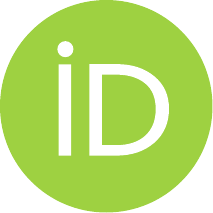}}%
}
\newcommand{\orcidlink}[1]{%
  \href{https://orcid.org/#1}{\orcidicon}%
}

\begin{document}

\title{QML-HCS: A Hypercausal Quantum Machine Learning Framework for Non-Stationary Environments\\
\thanks{}
}

\author{
\IEEEauthorblockN{
\href{https://orcid.org/0009-0008-2426-623X}{
Hector E Mozo \hspace{1mm}\includegraphics[scale=0.10]{orcid.pdf}}
}
\IEEEauthorblockA{
\textit{NeureonMindFlux Research Lab} \\
Jacksonville, Florida, United States \\
\texttt{hector.mozo@neureonmindfluxlab.org} \\
\texttt{hem10027@nyu.edu}
}
}

\maketitle
\bstctlcite{IEEEexample:BSTcontrol}

\begin{abstract}
QML-HCS is a research-grade framework for constructing and analyzing quantum-inspired machine learning models operating under hypercausal feedback dynamics. Hyper-causal refers to AI systems that leverage extended, deep, or non-linear causal relationships (expanded causality) to reason, predict, and infer states beyond the capabilities of traditional causal models. Current machine-learning and quantum-inspired systems struggle in non-stationary environments, where data distributions drift and models lack mechanisms for continuous adaptation, causal stability, and coherent state updating. QML-HCS addresses this limitation through a unified computational architecture that integrates quantum-inspired superposition principles, dynamic causal feedback, and deterministic–stochastic hybrid execution to enable adaptive behavior in changing environments.

The framework implements a hypercausal processing core capable of reversible transformations, multi-path causal propagation, and evaluation of alternative states under drift. Its architecture incorporates continuous feedback to preserve causal consistency and adjust model behavior without requiring full retraining. QML-HCS provides a reproducible and extensible Python interface backed by efficient computational routines, enabling experimentation in quantum-inspired learning, causal reasoning, and hybrid computation without the need for specialized hardware.

A minimal simulation demonstrates how a hypercausal model adapts to a sudden shift in the input distribution while preserving internal coherence. This initial release establishes the foundational architecture for future theoretical extensions, benchmarking studies, and integration with classical and quantum simulation platforms.
\end{abstract}


\section{Introduction}
Quantum-inspired machine learning has emerged as a promising paradigm that leverages principles such as superposition, parallel information flow, and reversible computation to improve learning efficiency and model expressiveness \cite{huynh2023qiml}. These approaches enable classical systems to mimic advantageous quantum-like behaviors without requiring quantum hardware. However, despite this progress, existing quantum-inspired and classical machine-learning models face significant challenges in non-stationary environments, where data distributions evolve over time and demand continuous causal coherence and adaptive stability. Traditional frameworks typically assume fixed causal structures or rely on retraining strategies that disrupt temporal continuity, making them ineffective in settings affected by concept drift \cite{hinder2023conceptdrift}, \cite{xuan2020bayesian}.

Hypercausal expands the capabilities of causal machine learning by incorporating extended, deep, and non-linear causal relationships that enhance the ability to reason, predict, and infer alternative system states. This broader causal foundation supports architectures capable of evaluating multiple causal pathways and maintaining model coherence as the environment changes. Recent work on causal discovery in non-stationary systems further illustrates the importance of models capable of restructuring causal dependencies as underlying conditions shift \cite{huang2019causal}.

QML-HCS is designed to address this gap by introducing a unified quantum-inspired hypercausal architecture tailored for adaptive learning and inference in evolving environments. The framework integrates superposition-inspired computation, dynamic causal feedback, and deterministic–stochastic hybrid execution to support stability and consistent model behavior without the need for complete retraining. Its Python interface enables reproducible experimentation, hybrid computation, and simulation of hypercausal structures on standard hardware.

This initial release focuses on establishing the foundational architecture required for hypercausal computation, including reversible processing pipelines, multi-path causal propagation, and continuous causal evaluation under drift. In addition to its native computational core, QML-HCS already provides adapters that interface with established external frameworks such as PennyLane and Qiskit, enabling the execution of quantum-inspired and hybrid workflows on widely used simulation backends. While this work emphasizes architectural clarity and minimal validation, future research will extend the system with comprehensive benchmarking, theoretical formalization, and integration with additional classical and quantum simulation platforms \cite{sellier2023quantum}. QML-HCS aims to provide a unified foundation for quantum-inspired and hypercausal processing suitable for adaptive learning systems operating in dynamic environments.

\section{Related Work}

Research in quantum-inspired and quantum-hybrid machine learning has produced several frameworks designed to enable efficient simulation of quantum-like computation on classical hardware. Among them, PennyLane offers a mature differentiable-programming interface for hybrid quantum–classical workflows and supports a variety of circuit-based quantum backends \cite{bergholm2022pennylane}. QML-HCS can optionally integrate PennyLane as an execution backend, benefiting from its established circuit engines. Nevertheless, although PennyLane incorporates some limited adaptive features, it remains fundamentally circuit-centric and does not provide the hypercausal feedback mechanisms, multi-path causal propagation, or drift-adaptive reasoning that QML-HCS introduces as part of its unified architecture. Qiskit offers a modular software stack for quantum circuit simulation and execution on IBM hardware, enabling circuit-based experimentation but lacking support for non-stationary or causal-adaptive architectures \cite{aleksandrowicz2019qiskit}. 

TensorFlow Quantum integrates quantum circuit layers into TensorFlow’s computational graph, providing tools for variational quantum models but focusing primarily on supervised learning tasks rather than dynamic causal adaptation \cite{broughton2021tensorflowquantum}. Qibo and Qadence extend quantum circuit simulation capabilities, with Qibo designed for high-performance classical simulation and Qadence provides a block-based, modular programming interface aimed at digital-analog quantum computing and supports primitives that are well-suited for quantum machine learning research \cite{efthymiou2024qibolab}, \cite{seitz2025qadence}. Although these frameworks facilitate quantum-inspired learning, none implement hypercausal feedback mechanisms, multi-path causal propagation, or adaptive reasoning under drift as in QML-HCS.

Within causal machine learning, several toolkits enable causal discovery and inference under controlled assumptions. DoWhy provides an extensible Python library for causal inference with automated identification and estimation routines \cite{sharma2020dowhy}. CausalML offers a suite of uplift modeling and treatment-effect estimation tools widely used in applied economics and policy evaluation \cite{chen2020causalml}. However, these frameworks treat causal structures as relatively static and do not incorporate dynamic causal models capable of adapting to evolving environments or enabling propagation through multiple causal branches. More recent works attempt to incorporate time-varying causal structures, such as temporal causal graphs and event-based causal modeling, but these approaches remain limited to offline estimation rather than continuous adaptive computation \cite{squires2023causalstructure}, \cite{malinsky2018causaltimeseries}. 

Efforts to address learning in non-stationary environments include frameworks for concept drift detection and dynamic model adaptation. River (formerly Creme) provides incremental learning tools for data streams and drift detection, enabling online updates but without support for quantum-inspired or causal branching mechanisms \cite{ergen2021convexgeometry}. Adaptive ensemble methods and drift-aware architectures have been proposed, yet they generally operate through retraining or model replacement rather than continuous causal feedback \cite{elwell2011incremental}. As a result, no existing framework offers a unified architecture that integrates quantum-inspired computation with hypercausal feedback mechanisms suited for dynamic environments. This gap motivates the development of QML-HCS.

\section{Framework Overview}

\begin{figure}[h]
    \centering
    \includegraphics[width=0.48\textwidth]{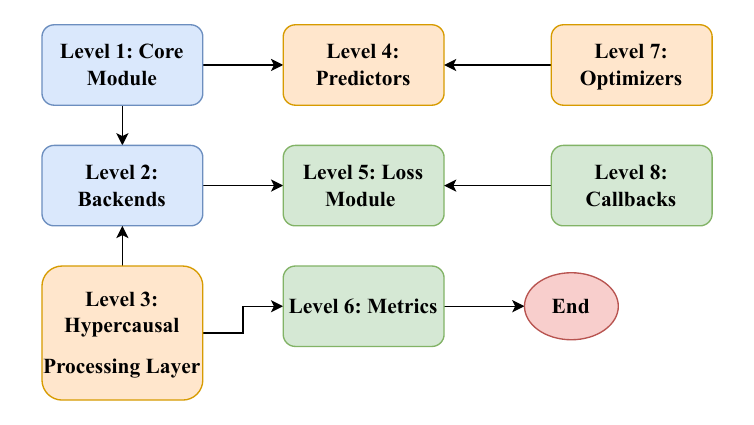}
    \caption{Layered architectural overview of QML-HCS. The figure illustrates the canonical structural organization of the system; it does not represent a required execution flow. It highlights eight conceptual layers: the core module, backend execution layer, hypercausal nodes, predictors, losses, metrics, optimizers, and callbacks.}
    \label{fig:qmlhcs-architecture}
\end{figure}

As previously introduced, QML-HCS is designed as a modular, research-grade system for constructing and analyzing quantum-inspired machine learning models that operate under hypercausal feedback dynamics. Its architecture follows a layered design philosophy that separates formal interface definitions, computational backends, causal propagation mechanisms, and evaluation modules, ensuring reproducibility, interoperability, and extensibility across heterogeneous environments.

At its foundation, the core layer defines typed protocols that govern how inputs, internal states, backend functions, and hypercausal nodes interact. This includes a triadic semantic flow expressed as input → state → projected futures, which serves as the unified execution model across the entire framework. A backend registry allows QML-HCS to dynamically instantiate analytic, stochastic, or compiled engines while maintaining mathematical and dimensional consistency.

The backend adapter layer provides the system with multiple execution environments. QML-HCS integrates analytic and differentiable execution through PennyLane, probabilistic sampled execution through Qiskit, and high-performance deterministic execution through compiled C++ backends exposed via pybind11. These adapters implement a shared interface, allowing users to switch execution modes without modifying model definitions.

The hypercausal layer defines the dynamic computational behavior of the framework. Hypercausal nodes encapsulate backend execution and generate multiple candidate futures, which are aggregated using selectable projection policies such as mean, median, or risk-minimizing estimators. This layer supports single-node computation, sequential processing chains, and full hypercausal graphs capable of handling drift, branching, and multi-step state propagation.

Evaluation and control are handled by the loss and metrics modules. These components quantify predictive error, coherence across branches, temporal consistency, anomaly likelihood, system stability, and alignment with target predictive behavior. Additional components, including the callbacks subsystem and the optimizer registry, provide telemetry logging, evaluation scheduling, and interchangeable gradient-based or gradient-free optimization strategies, enabling the system to adapt to a broad range of research workflows.

Together, these components form a unified pipeline in which user inputs pass through backend engines and hypercausal nodes, produce multi-branch future projections, undergo selection through projection policies, and are evaluated through task-specific, coherence-oriented, and stability-oriented metrics. The entire system operates from Python while optionally delegating computation to high-performance C++ engines, enabling rapid experimentation and scalable model execution.

\section{System Architecture}
\label{sec:system-architecture}
Building on the earlier conceptual overview, QML-HCS operates through a layered computational design structured across eight conceptual levels. Figure~\ref{fig:qmlhcs-architecture} introduced the canonical organization of these layers and their high-level relationships. In the present section, we formalize their internal architecture by detailing the mathematical foundations, execution semantics, state–transition mechanisms, and hypercausal propagation rules that govern each level from Level 1 through Level 8. This deeper exposition clarifies how the system performs computation, generates and evaluates hypercausal futures, and maintains coherence across analytic, stochastic, and compiled backend environments.

\subsection{Level 1: Core Module}

\begin{center}
\small
\begin{tabular}{p{1.2cm} p{6.0cm}}
\hline
\textbf{ID} & \textbf{Role in the Core Module} \\
\hline
{\small A: \texttt{types}} &
Defines core tensor aliases, state vectors $S_t$, future sets $F_t$, 
and shared protocol interfaces. \\[0.1cm]

{\small B: \texttt{backend}} &
Implements the backend execution contract: input encoding, state 
generation $S_t$, and future projection $F_t$. \\[0.1cm]

{\small C: \texttt{model}} &
Composes hypercausal nodes and applies the triadic rule 
$(x_t, S_{t-1}) \rightarrow (S_t, \hat S_{t+1})$ in single-node or 
chained configurations. \\[0.1cm]

{\small D: \texttt{registry}} &
Stores backend entries, capabilities, and provides a uniform 
backend instantiation interface. \\[0.1cm]

{\small E: \texttt{config}} &
Holds static configuration (output dimension, branch count, shots) 
that governs backend and model execution. \\
\hline
\end{tabular}
\end{center}

The Core Module establishes the fundamental execution contracts and typed 
semantics that structure all higher levels in QML-HCS. It defines the 
mathematical form of states, futures, backend operators, and triadic 
hypercausal propagation, ensuring that all components share a consistent 
computational interface.

\paragraph{State and Future Representations.}
The core defines numerical containers for current states and projected futures.  
A state vector is represented as
\[
S_t \in \mathbb{R}^{D},
\]
while a set of $K$ candidate futures is expressed as
\[
F_t \in \mathbb{R}^{K \times D},\qquad K \ge 2.
\]
These structures form the minimal interface for all backends and hypercausal
nodes, guaranteeing dimensional consistency throughout the framework.

\paragraph{Backend Execution Contract.}
Backends implement a three-step computational interface: input encoding,
state generation, and future projection. Given an input vector $x_t$ and a
configuration specifying output dimensionality $D$, a backend performs
\[
x_t \;\longrightarrow\; S_t = f(x_t),
\]
followed by the generation of $K$ projected futures,
\[
F_t = g(S_t, K).
\]
This contract ensures that analytic, stochastic, and compiled providers all
expose compatible behavior, regardless of their underlying execution model.

\paragraph{Triadic Hypercausal Propagation.}
Hypercausal nodes apply the core update rule used across the system.  
Given the input $x_t$ and optional previous state $S_{t-1}$, each node 
computes
\[
(x_t,\, S_{t-1}) \;\longrightarrow\; \big(S_t,\, \hat{S}_{t+1}\big),
\]
where $S_t$ is the current state and $\hat{S}_{t+1}$ is the aggregated
representative future chosen from $F_t$ according to a selection policy
(mean, median, or risk-minimizing).  
This mapping is the minimal semantic contract governing one-step hypercausal
processing.

\paragraph{Model Composition.}
At the system level, multiple hypercausal nodes may be composed sequentially.  
In a chain of nodes $\{N_1,\dots,N_M\}$, the output state of each node becomes
the input to the next:
\[
S_t^{(i)} = N_i\big(S_t^{(i-1)}\big),
\qquad S_t^{(0)} = x_t,
\]
with $\hat{S}_{t+1}$ generated by the final node in the chain.  
This ensures that multi-node compositions remain consistent with the core
execution semantics.

\paragraph{Backend Discovery and Instantiation.}
A lightweight registry maintains all available backend providers.  
Each backend is associated with a name, constructor, and static capability 
metadata.  
Given a configuration specifying the desired dimensionality and execution 
mode (analytic or shot-based), a backend instance is created through a uniform
factory interface, ensuring reproducibility and controlled initialization.

\paragraph{Functional Role.}
Through these components, the Core Module defines the minimal computational
building blocks of QML-HCS: the structure of states, the generation of futures,
the triadic propagation rule, the chaining semantics for multi-node execution,
and the unified mechanism for backend discovery and instantiation.  
All higher layers—predictors, metrics, losses, optimizers, and callbacks—are
constructed on top of these foundational contracts.

\subsection{Level 2: Backend Execution Layer}

\begin{center}
\small
\begin{tabular}{p{1.8cm} p{6.0cm}}
\hline
\textbf{ID} & \textbf{Execution Role} \\
\hline
\parbox{1.8cm}{\small A:\\[-2pt]\texttt{pennylane}} &
Analytic or differentiable execution using variational circuits; produces deterministic or shot-based state vectors. \\[0.12cm]

\parbox{1.8cm}{\small B:\\[-2pt]\texttt{qiskit}} &
Shot-based stochastic execution via the Sampler primitive; returns expectation-like vectors derived from measurement counts. \\[0.12cm]

\parbox{1.8cm}{\small C:\\[-2pt]\texttt{cpp}} &
Compiled deterministic execution with strict validation and high-performance state generation. \\
\hline
\end{tabular}
\end{center}

The backend execution layer provides QML-HCS with multiple quantum-inspired computation environments. Each adapter implements the unified backend interface defined in the Core Module, ensuring that all backends compute a state vector $S_t$ and a set of projected futures $F_t$ using the same semantic flow established previously. This allows model definitions to remain invariant across analytic, stochastic, and compiled execution modes.

\paragraph{PennyLane Adapter.}
The PennyLane backend constructs a lightweight variational circuit composed of RY rotations and linear entanglement. Execution can occur in analytic mode or using a finite number of shots. The circuit outputs Pauli-Z expectation values per wire, forming a state vector
\[
S_t[i] = \big\langle Z_i \big\rangle_{\,\rho(x_t)},
\]
where $\rho(x_t)$ denotes the quantum state prepared from the encoded input $x_t$ and $Z_i$ is the Pauli-Z operator on wire $i$. The resulting $S_t$ matches the output dimension specified by the backend configuration and remains compatible with the Level~1 state representation.

\paragraph{Qiskit Adapter.}
The Qiskit backend relies on the Sampler primitive to evaluate expectation-like values derived from bitstring frequency distributions. Let $c(b)$ be the count associated with bitstring $b \in \{0,1\}^D$ and
\[
N = \sum_{b} c(b)
\]
the total number of shots. After endian correction, the $i$th component of the state vector is computed as
\[
S_t[i] = \frac{1}{N} \sum_{b} c(b)\,\big(2\,b_i - 1\big),
\]
where $b_i \in \{0,1\}$ is the bit corresponding to wire $i$. This provides a stochastic, physically grounded counterpart to analytic evaluations while preserving the same dimensional structure.

\paragraph{C++ Adapter via pybind11.}
The C++ backend connects compiled quantum-inspired engines to Python through a minimal pybind11 bridge. The adapter validates the presence of the required functions, enforces consistency between the reported output dimension and the Python configuration, and executes deterministically. Given an encoded input $x_t$, the backend returns
\[
S_t = f_{\text{C++}}(x_t),
\]
where $f_{\text{C++}}$ denotes the compiled operator implemented in the underlying C++ engine.

\paragraph{Theoretical Foundations.}
The PennyLane adapter follows the standard formulation of variational quantum circuits (VQCs), where quantum states are prepared by parameterized unitary operations and evaluated through Pauli expectation values~\cite{schuld2019qml}. Conversely, the Qiskit backend implements sampling-based expectation estimation, in which observable values are reconstructed from bitstring measurement frequencies obtained over multiple shots~\cite{qiskit2025sampler}. These two paradigms, analytic variational evaluation and stochastic measurement-driven estimation, provide the theoretical grounding for the unified execution semantics adopted in QML-HCS.

\paragraph{Unified Execution Semantics.}
As we previously observed, despite differences in computational modality (analytic, sampled, or compiled), all adapters implement the same mapping from encoded inputs to states and projected futures as defined in Level~1. This guarantees that hypercausal nodes, projection policies, and downstream evaluators operate identically regardless of the chosen backend, enabling seamless switching between differentiable simulation, shot-based stochastic execution, and optimized compiled computation without modifying model structure or logic.

\subsection{Level 3: Hypercausal Layer}

\begin{center}
\small
\begin{tabular}{p{1.6cm} p{6.2cm}}
\hline
\textbf{ID} & \textbf{Functional Role} \\
\hline
\parbox{1.6cm}{\small A:\\[-2pt]\texttt{HCNode}} &
Encapsulates backend execution; generates $K$ candidate futures and applies a projection policy to obtain the representative future. \\[0.12cm]

\parbox{1.6cm}{\small B:\\[-2pt]\texttt{Policies}} &
Aggregation strategies (mean, median, min-risk) that select a single future from the $K$ candidates. \\[0.12cm]

\parbox{1.6cm}{\small C:\\[-2pt]\texttt{HCGraph}} &
Executes hypercausal nodes in topological order, supporting chains and DAGs with automatic parent-state averaging. \\
\hline
\end{tabular}
\end{center}

The hypercausal layer defines the dynamic computational behavior of QML-HCS. Hypercausal nodes wrap backend execution and generate multiple candidate futures, which are aggregated through selectable projection policies. This layer supports single-node computation, sequential chains, and full hypercausal graphs capable of modeling drift, branching, and multi-step propagation.

\paragraph{Hypercausal Node.}
Each node binds a backend and an optional projection policy. Given the current input and the optional past state, the node computes the present state and a set of $K$ future candidates. The node then selects a representative future using its assigned policy. Formally, if $F$ denotes the $(K \times D)$ matrix of candidate futures, a policy computes
\[
\hat{S}_{t+1} = \pi(F),
\]
where $\pi$ is the aggregation rule (mean, median, or min-risk). If no explicit policy is provided, the element-wise mean is used.

\paragraph{Projection Policies.}
The policies determine how $K$ branches are reduced to a single representative vector:
\[
\text{Mean:}\quad \pi_{\mathrm{mean}}(F) = \mathrm{mean}(F, 0),
\]
\[
\text{Median:}\quad \pi_{\mathrm{med}}(F) = \mathrm{median}(F, 0),
\]
\[
\text{Min-risk:}\quad \pi_{\mathrm{risk}}(F) = F_{i^\ast}, \qquad
i^\ast = \arg\min_i r(F_i),
\]
where $r(\cdot)$ is a user-defined risk functional. These policies provide different behaviors, allowing the hypercausal layer to emphasize stability, robustness, or conservative forecasting depending on the task.

\paragraph{Architectural Context.}
Although QML-HCS is not based on continuous-time formulations, its state-transition mechanism is conceptually related to modern dynamical models such as Neural Ordinary Differential Equations (Neural ODEs)~\cite{chen2019neuralode}, which describe how internal states evolve through learned differential flows. In contrast, QML-HCS implements a discrete hypercausal transition that produces multiple candidate futures and selects a representative one via projection policies, extending classical dynamical modeling with multi-branch, drift-adaptive computation.

\paragraph{Hypercausal Graphs.}
The graph module provides deterministic evaluation of hypercausal nodes in directed acyclic topologies. Nodes are executed in topological order. If a node lacks an explicit input, its input is derived as the mean of its parent states:
\[
x_t = \frac{1}{|P|}\sum_{p \in P} S_t^{(p)},
\]
where $P$ denotes the set of parent nodes. The graph returns per-node present states, projected futures, and diagnostics. Both linear chains and general DAGs are supported.

\paragraph{Clarification.}
Although the hypercausal layer employs directed acyclic graphs to organize internal computation, these DAGs are \emph{not} structural causal models in the Pearl--Sch{\"o}lkopf sense. Classical causal DAGs encode real-world cause--effect mechanisms and support intervention semantics within structural causal models (SCMs)~\cite{scholkopf2021causalrepresentationlearning}, whereas QML-HCS uses DAGs solely to coordinate state propagation, multi-branch future generation, and backend-driven hypercausal dynamics within its computational architecture.

\paragraph{Functional Role.}
Through nodes, policies, and graph evaluation, the hypercausal layer provides the mechanisms required to construct dynamic multi-branch computation, assemble sequential or branched structures, and propagate state information across heterogeneous computational paths. This layer enables higher-level components-predictors, evaluators, and optimizers-to operate over consistent, policy-driven hypercausal dynamics.

\subsection{Level 4: Predictors and Projectors}

\begin{center}
\small
\begin{tabular}{p{1.6cm} p{6.2cm}}
\hline
\textbf{ID} & \textbf{Functional Role} \\
\hline
\parbox{1.6cm}{\small A:\\[-2pt]\texttt{Projector}} &
Defines the contract for deterministic future projection from a compact state; produces $K$ candidate futures. \\[0.12cm]

\parbox{1.6cm}{\small B:\\[-2pt]
  \texttt{Linear\\Projector}
} &
Implements affine prediction followed by evenly spaced perturbations and $\tanh$ stabilization to generate structured future sets. \\[0.12cm]

\parbox{1.6cm}{\small C:\\[-2pt]\texttt{Anticipator}} &
Builds counterfactual futures by augmenting baseline projector outputs with perturbation-driven variants and optional mirror symmetry. \\
\hline
\end{tabular}
\end{center}

The predictor and projector layer provides QML-HCS with deterministic future–generation mechanisms. Projectors map the compact present state into a multi-branch set of candidate futures, while anticipators extend these projections by injecting structured counterfactual variants. Together, these components supply the hypercausal layer with the multi-future tensors required for downstream aggregation, evaluation, and sequential propagation.

\paragraph{Deterministic Projectors.}
A projector implements the contract
\[
F = \Pi(s_t, K),
\]
where $\Pi$ denotes the projection operator and $F$ is a $(K \times D)$ future matrix.  
This abstraction ensures compatibility with the hypercausal layer without imposing assumptions on how the futures are generated.

\paragraph{Linear Projector.}
The \texttt{LinearProjector} constructs structured futures using an affine base prediction followed by evenly spaced perturbations.  
Given a compact state $s_t$, the projector computes
\[
\text{base} = w\,s_t + b,
\]
and generates $K$ deltas uniformly in the range $[-\mathrm{span}, \mathrm{span}]$.  
Each candidate future is then
\[
F_k = \tanh\!\big(\text{base} + \delta_k\big),
\]
which provides numerical stability and preserves the configured output dimension.  
This mechanism yields smooth, ordered future variations suited for deterministic exploration.

\paragraph{Contrafactual Anticipator.}
The anticipator augments any base projector with perturbation-driven counterfactuals.  
Given the base future set $F$ from a projector,
\[
F = \Pi(s_t, K),
\]
the anticipator computes its center
\[
c = \mathrm{mean}(F, 0),
\]
and, for each perturbation $p$, appends the variant
\[
v = p(c),
\]
and, if symmetric mode is enabled,
\[
v_{\mathrm{mir}} = 2c - v.
\]
The final future matrix is the concatenation of the original projector outputs and all perturbation-based variants.  
This enables the system to simulate structured deviations, mirrored scenarios, and counterfactual trajectories around the baseline prediction.

\paragraph{Functional Role.}
Through deterministic projection and counterfactual augmentation, this layer defines how QML-HCS constructs its multi-branch futures prior to policy selection. These mechanisms provide the raw candidate futures upon which hypercausal nodes, aggregation policies, and graph-level propagation operate, enabling the system to explore both direct predictions and hypothetical alternatives within a unified computational interface.

\subsection{Level 5: Losses and Metrics}

\begin{center}
\small
\begin{tabular}{p{1.6cm} p{6.2cm}}
\hline
\textbf{ID} & \textbf{Functional Role} \\
\hline

\parbox{1.6cm}{
  \small A:\\[-2pt]
  {\ttfamily Coherence\\Loss}
} &
Penalizes dispersion across future branches using variance or mean absolute deviation. \\[0.12cm]

\parbox{1.6cm}{
  \scriptsize B:\\[-2pt]
  {\ttfamily Consistency\\Loss}
} &
Enforces temporal smoothness across the triad $(S_{t-1}, S_t, \hat{S}_{t+1})$ through weighted deviations. \\[0.12cm]

\parbox{1.6cm}{
  \small C:\\[-2pt]
  {\ttfamily Task\\Losses}
} &
Provides standard task objectives including MSE, MAE, and cross-entropy for prediction accuracy. \\

\hline
\end{tabular}
\end{center}

Evaluation and control within QML-HCS rely on losses that quantify dispersion, temporal coherence, and task-specific predictive accuracy. These components provide the optimization signals that stabilize hypercausal propagation and align model behavior with supervised targets.

\paragraph{Inter-Branch Coherence Loss.}
This loss penalizes spread among the $K$ futures $F \in \mathbb{R}^{K \times D}$.  
Let $\mu = \mathrm{mean}(F, 0)$ be the per-dimension center.  
Two dispersion modes are provided:
\[
L_{\mathrm{var}} = \frac{1}{KD}\sum_{k=1}^{K}\sum_{j=1}^{D}\big(F_{k,j} - \mu_j\big)^2,
\]
\[
L_{\mathrm{mad}} = \frac{1}{KD}\sum_{k=1}^{K}\sum_{j=1}^{D}\left|F_{k,j} - \mu_j\right|.
\]
Variance encourages smooth dispersion, while MAD provides robustness to outliers.

\paragraph{Triadic Consistency Loss.}
Given states $S_{t-1}$, $S_t$, and predicted $\hat{S}_{t+1}$, temporal alignment is encouraged by
\[
L_{\mathrm{tri}} =
\alpha\,\|S_t - S_{t-1}\|^{2}
\;+\;
\beta\,\|S_t - \hat{S}_{t+1}\|^{2},
\]
with $\alpha$ and $\beta$ governing sensitivity to deviations from past and predicted transitions.

\paragraph{Task-Level Predictive Losses.}
Task losses measure agreement between predictions $p$ and targets $t$.  
Three objectives are supported:
\[
L_{\mathrm{MSE}} = \frac{1}{N}\sum_{i=1}^{N}(p_i - t_i)^2,
\]
\[
L_{\mathrm{MAE}} = \frac{1}{N}\sum_{i=1}^{N}|p_i - t_i|,
\]
\[
L_{\mathrm{CE}} = -\sum_{i=1}^{N} t_i \log\!\left(\frac{p_i}{\sum_j p_j}\right).
\]
Inputs are flattened and clipped for numerical stability.

\paragraph{Functional Role.}
Together, coherence, consistency, and task losses regulate dispersion, enforce temporal structure, and align predictions with supervised objectives. These modules supply the numerical gradients or evaluation signals driving stable and interpretable hypercausal learning.

\subsection{Level 6: Anomalies, Control, and Forecasting Metrics}

\begin{center}
\small
\begin{tabular}{p{1.8cm} p{6.2cm}}
\hline
\textbf{ID} & \textbf{Functional Role} \\
\hline

\parbox{1.8cm}{
  \small A:\\[-2pt]
  {\ttfamily Anomaly\\Metrics}
} &
Detect deviations from expected hypercausal behavior using ROC-based detection and lag-aware recall. \\[0.15cm]

\parbox{1.8cm}{
  \small B:\\[-2pt]
  {\ttfamily Control\\Metrics}
} &
Quantify dynamical stability through overshoot, settling time, and robustness indicators for state response. \\[0.15cm]

\parbox{1.8cm}{
  \small C:\\[-2pt]
  {\ttfamily Forecasting\\Metrics}
} &
Evaluate predictive accuracy using MAPE, MASE, $\Delta$-lag error, and RMSE for multi-step forecasting. \\
\hline
\end{tabular}
\end{center}

The sixth level of QML-HCS expands evaluation beyond task accuracy to include anomaly detection, dynamical control diagnostics, and forecasting precision. These metrics enable the framework to characterize stability, detect regime shifts, and quantify predictive quality within hypercausal environments.

\paragraph{A. Anomaly Metrics.}
Anomaly metrics measure deviations between predicted structure and observed behavior.  
The ROC-based detector computes
\[
\mathrm{ROC\_AUC}
=
\frac{1}{2}
\left(
  \frac{\sum_{i} \mathrm{TPR}_i}{N}
  +
  \frac{\sum_{i} \mathrm{FPR}_i}{N}
\right),
\]
while lag-aware recall focuses on anomaly persistence over $\ell$ steps:
\[
\mathrm{Recall}_{\ell}
=
\frac{1}{M}
\sum_{t=1}^{M}
\mathbb{I}\big(\exists\, k \le \ell:\; y_{t+k}=1\big).
\]
These provide sensitivity to both instantaneous and temporally extended anomalies.

\paragraph{B. Control Metrics.}
Control-oriented metrics evaluate dynamical response properties in the triad $(S_{t-1}, S_{t}, \hat{S}_{t+1})$.  
Overshoot is defined as
\[
\mathrm{OS}
=
\frac{\max_j S_{t,j} - S_{t-1,j}}{S_{t-1,j}},
\]
while settling time measures the duration until the deviation stays within a tolerance $\epsilon$:
\[
T_{\mathrm{settle}}
=
\min\{k:\; \|S_{t+k} - S_{t}\| < \epsilon\}.
\]
Robustness is evaluated through bounded-response criteria under perturbations.

\paragraph{C. Forecasting Metrics.}
Forecasting metrics quantify predictive accuracy across multi-step or drift-prone scenarios.  
Mean absolute percentage error is
\[
\mathrm{MAPE}
=
\frac{100}{N}
\sum_{i=1}^{N}
\left|
\frac{p_i - t_i}{t_i}
\right|,
\]
while MASE normalizes error by a naive one-step baseline:
\[
\mathrm{MASE}
=
\frac{
  \frac{1}{N}\sum_{i=1}^{N}|p_i - t_i|
}{
  \frac{1}{N-1}\sum_{i=2}^{N}|t_i - t_{i-1}|
}.
\]
Lag-delta error captures horizon misalignment:
\[
\Delta_\ell
=
\frac{1}{N}\sum_{i=1}^{N} |p_{i+\ell} - t_i|,
\]
and RMSE evaluates Euclidean deviation:
\[
\mathrm{RMSE}
=
\sqrt{
  \frac{1}{N}\sum_{i=1}^{N}(p_i - t_i)^2
}.
\]

\paragraph{Functional Role.}
Taken together, these metrics characterize anomaly likelihood, dynamical stability, and predictive precision in hypercausal systems. They supply essential diagnostic signals for monitoring performance, identifying drift, and ensuring robust state evolution across complex environments.

\subsection{Level 7: Optimizers}

\begin{center}
\small
\begin{tabular}{p{1.8cm} p{6.2cm}}
\hline
\textbf{ID} & \textbf{Functional Role} \\
\hline

\parbox{1.8cm}{
  \small A:\\[-2pt]
  {\ttfamily Optimizer\\API}
} &
Defines a unified interface for parameter updates, exposing initialization,
state handling, and update semantics compatible with both internal
and external optimization engines. \\[0.15cm]

\parbox{1.8cm}{
  \small B:\\[-2pt]
  {\ttfamily NumPy\\Registry}
} &
Provides built-in optimizer constructors and supports external algorithms
such as adaptive, gradient-based, gradient-free, and hybrid optimization
methods through a dynamic factory mechanism. \\
\hline
\end{tabular}
\end{center}

The optimization layer standardizes how parameter updates are performed 
through a single programmable abstraction. This design separates the logic 
of parameter modification from the computational backends, enabling 
hypercausal predictors and analytic or compiled engines to share a consistent 
update mechanism.

\paragraph{A. Unified Optimizer API.}
The Optimizer API defines the contract for all optimization routines used 
within QML-HCS. It specifies methods for creating optimizer state, updating 
parameters, and integrating with both differentiable and non-differentiable 
execution modes.  
Because the API is fully modular, it supports internal optimizers as well as 
external libraries, including NumPy-based methods, PyTorch-style optimizers, 
C++ compiled routines, evolutionary strategies, and custom research-grade 
update rules.

\paragraph{B. NumPy Optimizer Registry.}
The NumPy registry serves as a dynamic factory that maps string identifiers 
to optimizer implementations. It supports a broad family of optimization 
strategies, including:
\begin{itemize}
    \item gradient-based methods (e.g., SGD-type updates),
    \item adaptive algorithms (Adam-like or moment-based),
    \item gradient-free methods (finite-difference, SPSA-like),
    \item hybrid or meta-heuristic approaches (evolutionary or sampling-based),
    \item externally provided optimizers registered by the user.
\end{itemize}

Because the registry accepts external entries, users can introduce 
custom optimizers, connect to compiled C++ update engines, replace internal 
methods with high-performance variants, or integrate existing optimization 
libraries without modifying model definitions.

\paragraph{Functional Role.}
Together, the Optimizer API and Registry form a flexible optimization layer 
capable of supporting research workflows that rely on analytic gradients, 
stochastic estimates, adaptive moment updates, or fully gradient-free 
strategies. This ensures that hypercausal learning remains effective across 
analytic, sampled, or compiled execution environments while remaining 
interoperable with external optimization ecosystems.

\subsection{Level 8: Callbacks and Telemetry}

\begin{center}
\small
\begin{tabular}{p{1.8cm} p{6.2cm}}
\hline
\textbf{ID} & \textbf{Functional Role} \\
\hline

\parbox{1.8cm}{
  \small A:\\[-2pt]
  {\ttfamily Callback\\Base}
} &
Provides an abstract event interface for coordinating auxiliary behaviors
during training, evaluation, and monitoring. \\[0.15cm]

\parbox{1.8cm}{
  \small B:\\[-2pt]
  {\ttfamily Depth\\Scheduler}
} &
Applies a mathematically defined schedule for adjusting a depth-like
parameter through linear interpolation across epochs. \\[0.15cm]

\parbox{1.8cm}{
  \small C:\\[-2pt]
  {\ttfamily Telemetry\\Loggers}
} &
Generates structured records capturing the evolution of system states and
events for analysis, debugging, and experiment reproducibility. \\
\hline
\end{tabular}
\end{center}

This layer introduces adaptive scheduling and structured runtime monitoring
mechanisms that operate independently of the model’s predictive and
optimization components. It supports dynamic behavior, capacity adaptation,
and transparent recording of system evolution.

\paragraph{A. Callback Interface.}
Callbacks define a conceptual mechanism for injecting auxiliary logic at key
moments of computation, such as the beginning or end of iterative steps,
epochs, or error events.  
They operate through a shared context that exposes high-level state
information-without modifying the mathematical semantics of prediction,
future generation, or optimization.  
Multiple callbacks may be composed, ensuring that monitoring, scheduling,
and control logic coexist without altering the core framework.

\paragraph{B. Depth Scheduling.}
The depth scheduler adjusts a depth-like model or backend parameter over
training epochs using a continuous interpolation rule.  
Given initial and final depth values,
$\mathrm{start}$ and $\mathrm{end}$, and a training horizon of $E$ epochs,
the depth assigned at epoch $e$ is
\[
\mathrm{depth}(e)
=
\mathrm{round}\!\Big(
  \mathrm{start}
  +
  (\mathrm{end} - \mathrm{start})
  \cdot
  \mathrm{clip}(e/E,\, 0,\, 1)
\Big).
\]
This interpolation ensures a smooth, bounded progression of model capacity,
allowing depth to increase gradually in alignment with stability and
learning requirements.

\paragraph{C. Telemetry Logging.}
Telemetry components generate structured records of system behavior,
capturing the temporal evolution of events, internal states, and selected
metadata.  
Each record consists of:
\[
r = (\tau,\, \sigma,\, \xi),
\]
where  
\(\tau \in \mathbb{R}\) is a timestamp,  
\(\sigma\) is an event label,  
and \(\xi\) contains contextual quantities relevant to the event.  
These records may be stored in in-memory buffers for rapid inspection or
serialized into log files for long-term monitoring, reproducibility, and
post-hoc analysis.

\paragraph{Functional Role.}
Callbacks and telemetry provide structural support for adaptive behaviors
and transparent monitoring within QML-HCS.  
They orchestrate runtime-side processes—such as scheduling, logging, and
control—while preserving the mathematical integrity of prediction,
hypercausal propagation, and optimization.

\section{Examples and Validation}
\label{sec:examples}

This section illustrates how QML-HCS behaves under hardware-style drift and
hypercausal feedback. We consider a single-scalar control parameter
$\alpha_t$ acting on a fixed input pattern, while the environment introduces
phase detuning, amplitude drift, and readout bias over $T$ epochs. The goal
is to evaluate whether the framework maintains stable hypercausal behavior
and metrically coherent dynamics in the presence of structured
non-stationarity.

\subsection{Experimental Configuration}

We instantiate a hypercausal model with a state dimension $D=7$, 
$K=20$ branches, and $T=300$ epochs. The base input profile
$x_0 \in \mathbb{R}^{D}$ is a fixed ramp, and a scalar feedback parameter
$\alpha_t$ rescales it at each epoch:
\[
x_t = \alpha_t x_0 .
\]
The model produces a current state $S_t$ and a representative future
$\hat{S}_{t+1}$ according to the hypercausal propagation rule described in
Section~\ref{sec:system-architecture}. Task, consistency, and coherence
losses are combined into a single scalar objective,
\[
L_t = L_{\mathrm{task},t}
    + \tfrac{1}{2}\big(
        L_{\mathrm{cons},t}
        + L_{\mathrm{coh},t}
      \big),
\]
which is minimized with respect to $\alpha_t$ by a stochastic,
trust-region-style optimizer.

The physical drift model emulates three hardware effects: phase drift,
frequency detuning, and readout bias. Let $t \in \{0,\dots,T-1\}$ denote the
epoch index. The phase drift applied additively to the input is
\[
\phi_t
=
\phi_{\max}
\sin\!\Big(
  \frac{2\pi t}{T-1}
\Big),
\]
while a slow multiplicative detuning acts as
\[
a_t = 1 + \varepsilon\, t,
\]
with $\varepsilon$ on the order of tens of parts per million. An oscillatory
readout bias is modeled as
\[
b_t
=
b_{\max}
\Big(
  \tfrac{1}{2}
  +
  \tfrac{1}{2}
  \sin(\phi_t + \varphi_0)
\Big),
\]
where $\varphi_0$ is a fixed phase offset. The model input and observed
state thus become
\[
\tilde{x}_t = a_t \big(x_t + \phi_t\big),
\qquad
S_t^{\mathrm{obs}}
=
(1 - b_t) S_t + b_t\, \mathrm{sign}(S_t),
\]
and the same post-processing is applied to $\hat{S}_{t+1}$. This setting
forces the hypercausal system to adapt to non-stationarity arising from
both pre-measurement and post-measurement perturbations.

\begin{center}
\small
\begin{tabular}{p{2.4cm} p{5.6cm}}
\hline
\textbf{Quantity} & \textbf{Configuration} \\
\hline
State dimension $D$     & $7$ \\
Branches $K$            & $20$ \\
Epochs $T$              & $300$ \\
Shots per evaluation    & $1024$ \\
Depth schedule          & linear from $1$ to $5$ across $T$ \\
Control parameter       & scalar feedback $\alpha_t$ \\
Drift model             & sinusoidal phase, linear detuning, oscillatory bias \\
Loss                    & task + consistency + coherence aggregate $L_t$ \\
\hline
\end{tabular}
\end{center}

\subsection{Drift-Aware Hypercausal Training}

\begin{figure}[H]
  \centering
  \includegraphics[width=0.42\textwidth]{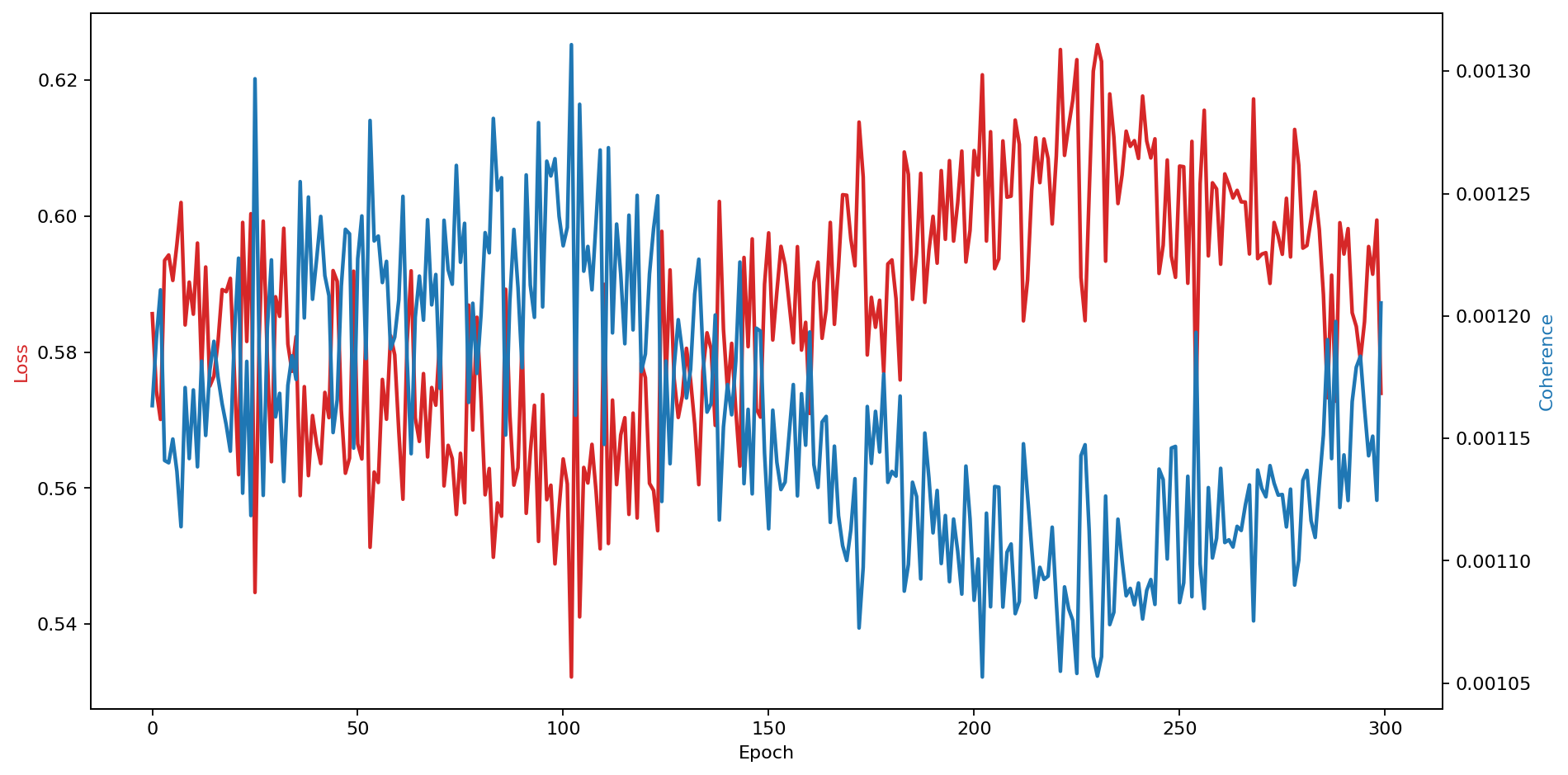}
  \caption{Aggregate loss and coherence component over epochs under hardware-style drift.}
  \label{fig:loss-coh}
\end{figure}

Figure~\ref{fig:loss-coh} shows the aggregate loss $L_t$ and the coherence
component across training. Both remain bounded and stable despite the
injected physical drift, indicating that the hypercausal pipeline can
maintain predictive performance and branch coherence under non-stationary
conditions. The depth schedule gradually increases model capacity yet does
not induce instability in the loss trajectory.

\begin{figure}[H]
  \centering
  \includegraphics[width=0.48\textwidth]{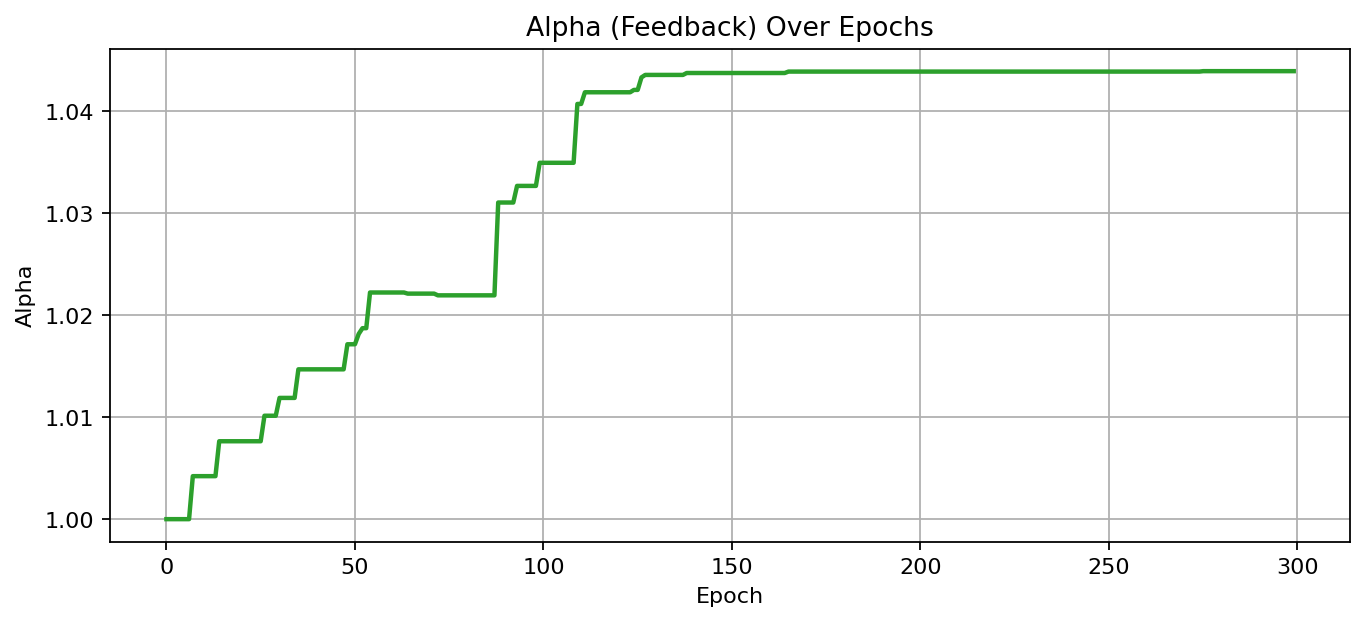}
  \caption{Evolution of the feedback parameter $\alpha_t$ across epochs.}
  \label{fig:alpha-over-epochs}
\end{figure}

\begin{figure}[H]
  \centering
  \includegraphics[width=0.48\textwidth]{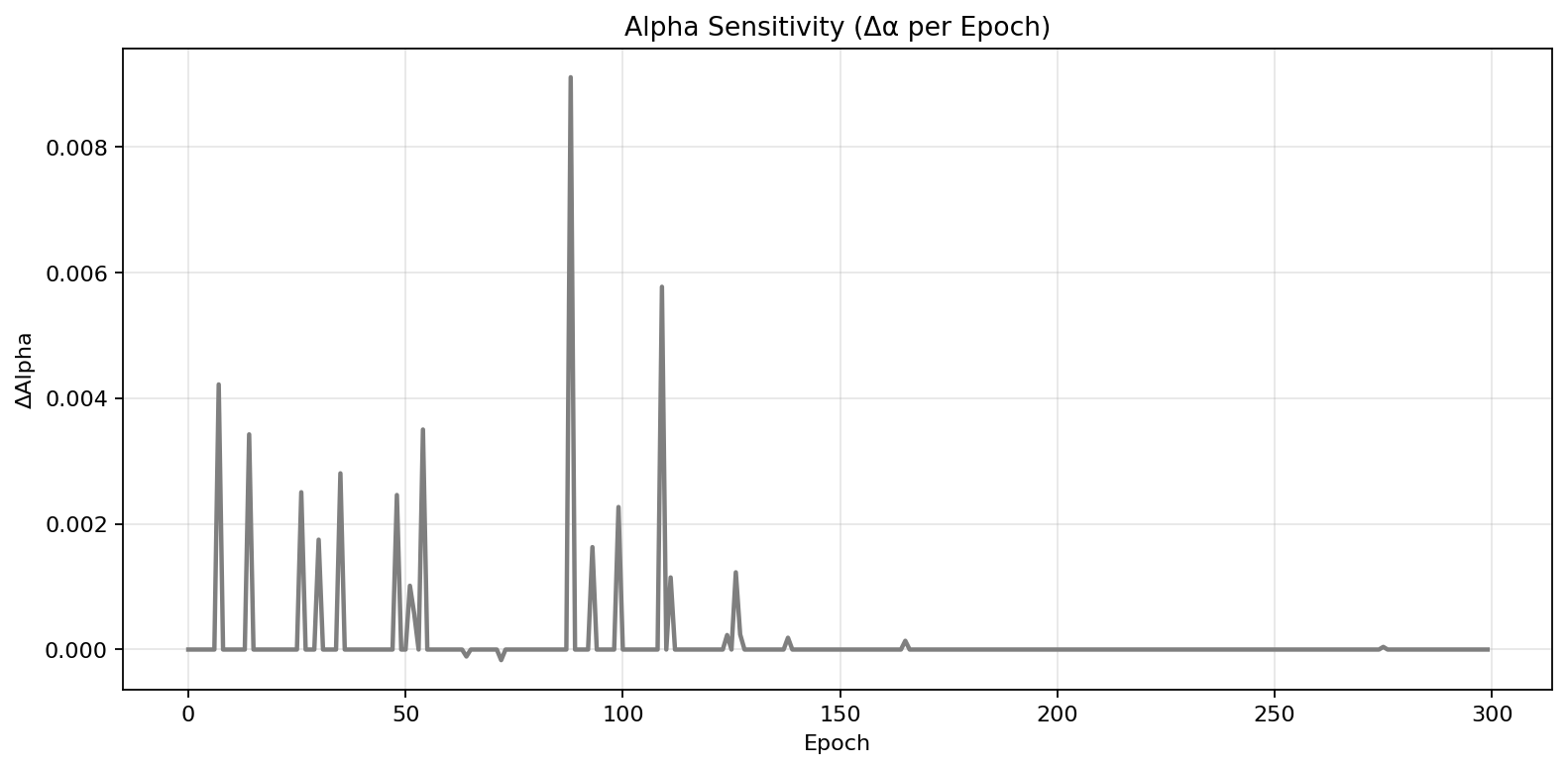}
  \caption{Per-epoch sensitivity $\Delta\alpha_t$ illustrating early adaptation and stabilization.}
  \label{fig:alpha-sensitivity}
\end{figure}

The feedback parameter $\alpha_t$ and its sensitivity are displayed in
Figures~\ref{fig:alpha-over-epochs} and~\ref{fig:alpha-sensitivity}.  
The trajectory of $\alpha_t$ exhibits a monotonic increase from the initial
value $\alpha_0 = 1.0$ to a plateau around $1.04$, after which only minor
adjustments occur. The per-epoch increment
\[
\Delta\alpha_t = \alpha_t - \alpha_{t-1}
\]
shows large corrections in the early epochs, followed by rapidly decaying
fluctuations that converge towards zero.


To summarize how states and projected futures remain aligned, we report the
mean value of the current state and the representative future,
\[
\bar{S}_t = \frac{1}{D}\sum_{i=1}^{D} S_{t,i},
\qquad
\bar{\mu}_t = \frac{1}{D}\sum_{i=1}^{D} \hat{S}_{t+1,i}.
\]


\begin{figure}[H]
  \centering
  \includegraphics[width=0.48\textwidth]{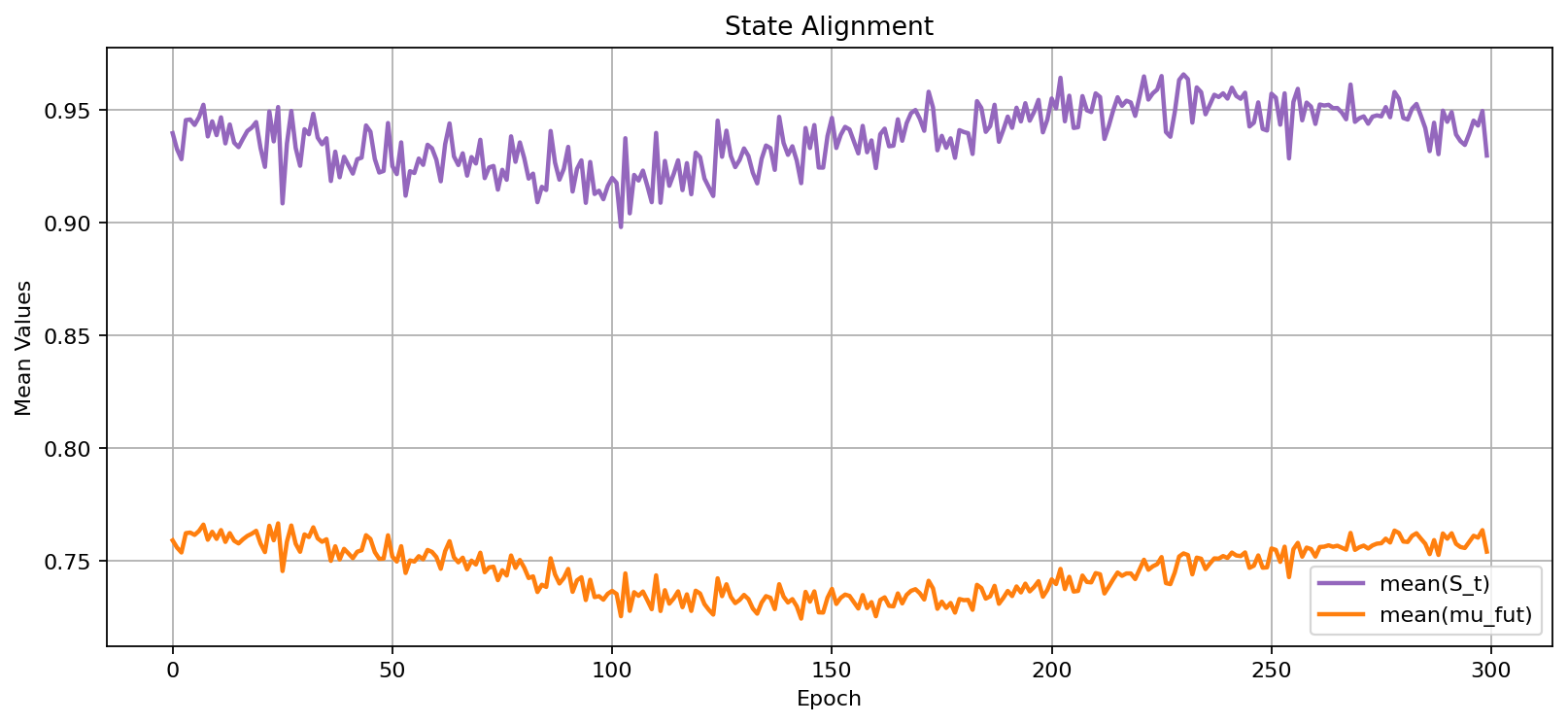}
  \caption{State alignment between mean state $\bar S_t$ and mean projected future $\bar{\mu}_t$.}
  \label{fig:state-alignment}
\end{figure}

Both curves occupy narrow bands, indicating that the hypercausal projection
policy preserves a stable relationship between present and future summaries
even in the presence of drift and readout perturbations.

\subsection{Causal Metric Geometry}

\begin{figure}[H]
  \centering
  \includegraphics[width=0.48\textwidth]{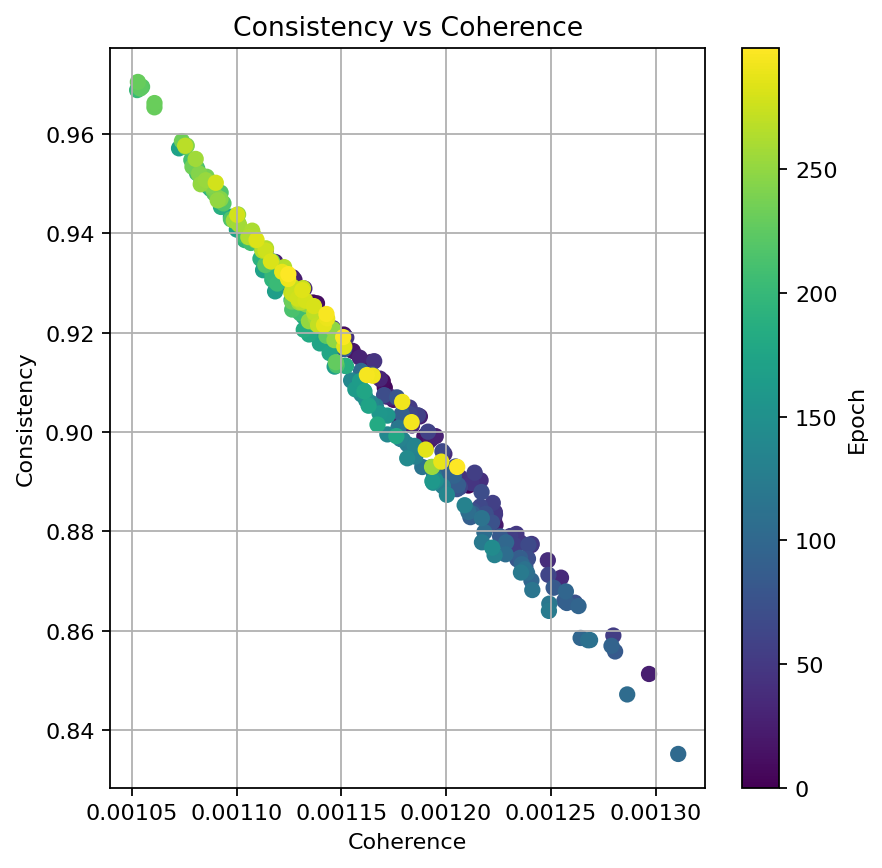}
  \caption{Consistency versus coherence across epochs.}
  \label{fig:consistency-vs-coherence}
\end{figure}

\begin{figure}[H]
  \centering
  \includegraphics[width=0.48\textwidth]{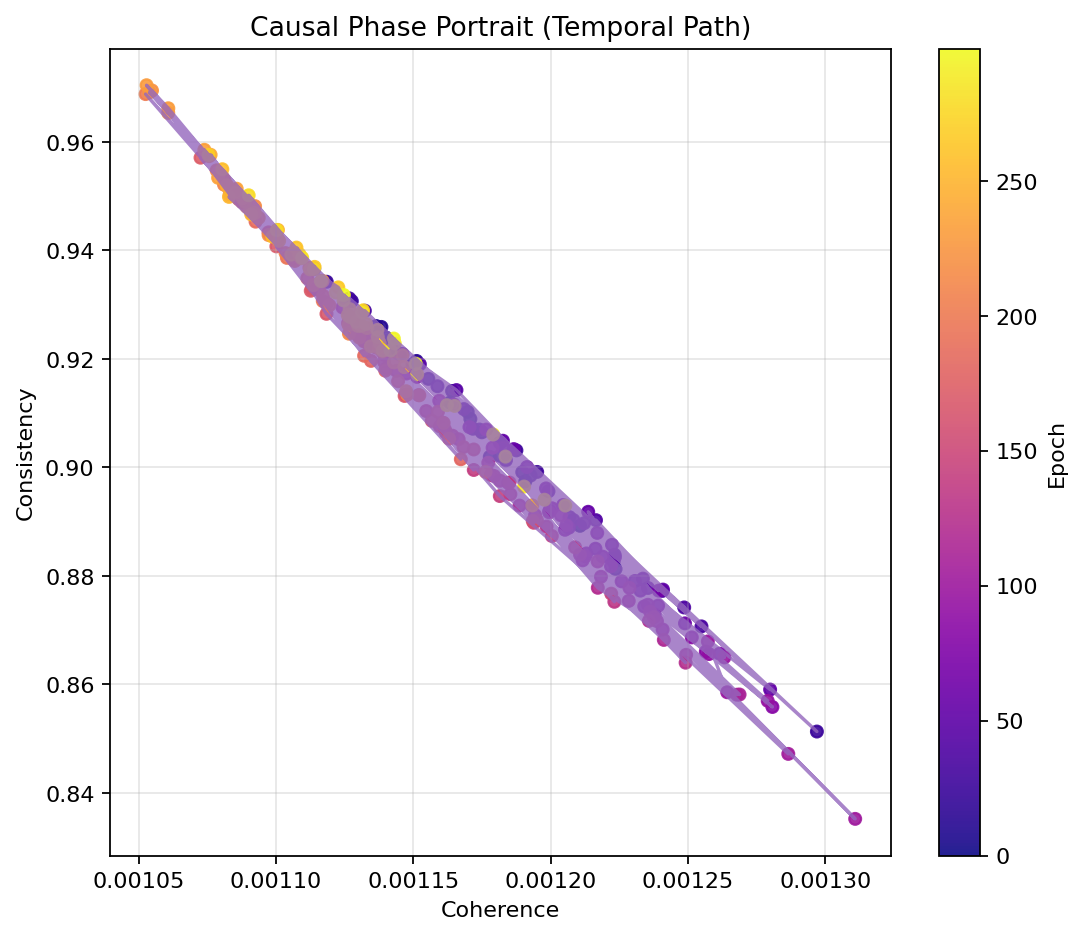}
  \caption{Causal phase portrait showing the temporal trajectory in the metric plane.}
  \label{fig:causal-phase}
\end{figure}

The joint evolution of coherence and consistency metrics is summarized in
Figure~\ref{fig:consistency-vs-coherence} and the causal phase portrait in
Figure~\ref{fig:causal-phase}.


The points form a narrow, negatively sloped band in the
$(L_{\mathrm{coh}}, L_{\mathrm{cons}})$ plane, revealing a nearly
one-dimensional trade-off manifold.

\begin{figure}[H]
  \centering
  \includegraphics[width=0.48\textwidth]{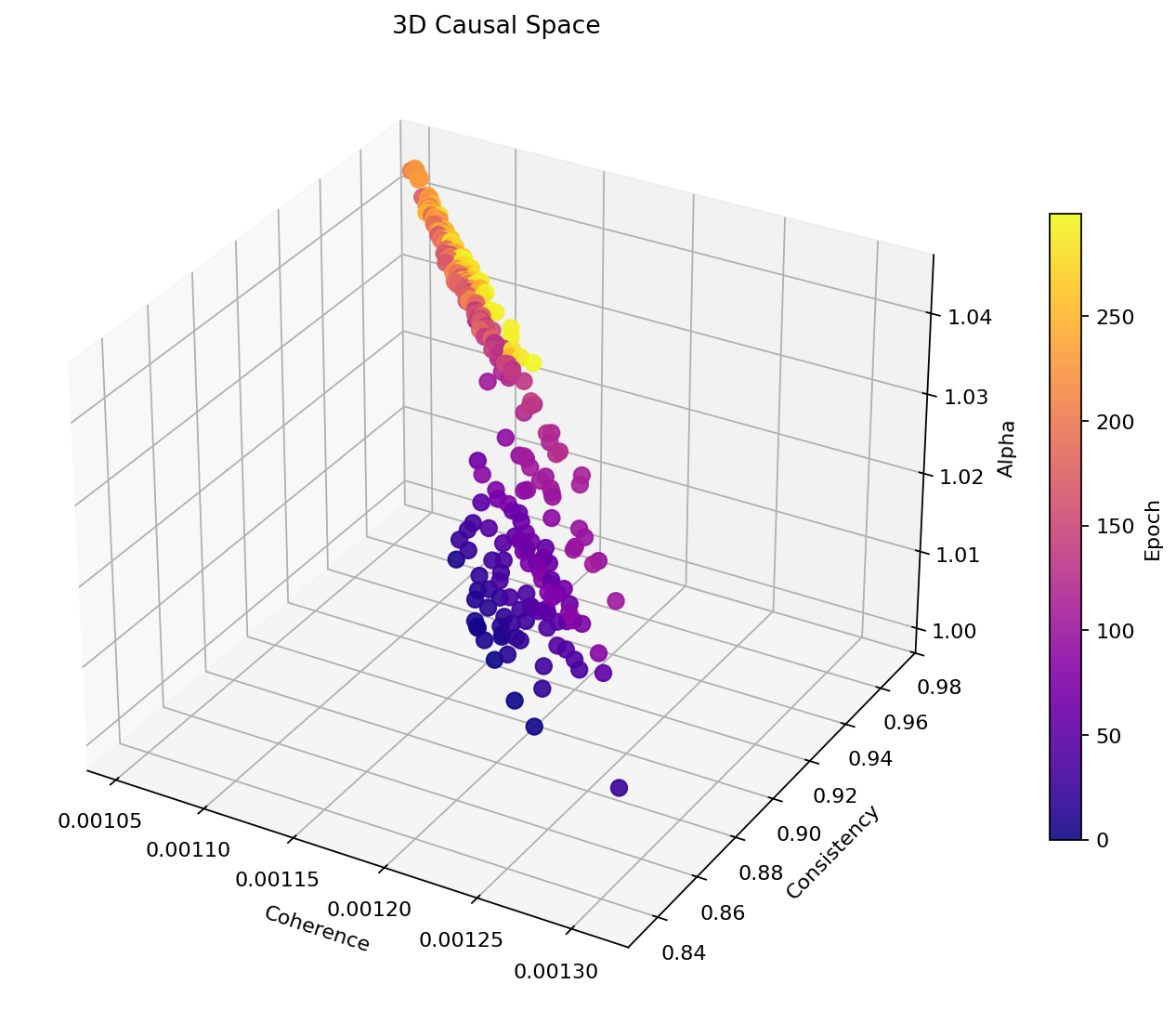}
  \caption{Three-dimensional causal space $(L_{\mathrm{coh}},L_{\mathrm{cons}},\alpha_t)$ colored by epoch.}
  \label{fig:causal-space-3d}
\end{figure}

The embedding shows that $\alpha_t$ acts as a low-dimensional control knob
that reconfigures hypercausal dynamics without destabilizing consistency or
coherence.

\subsection{Drift Proxies and Feedback Sensitivity}

To assess whether the framework exposes drift-relevant signals, we treat
$\Delta\alpha_t$ as a drift proxy and compare it with the known phase drift
$\phi_t$.


\begin{figure}[H]
  \centering
  \includegraphics[width=0.48\textwidth]{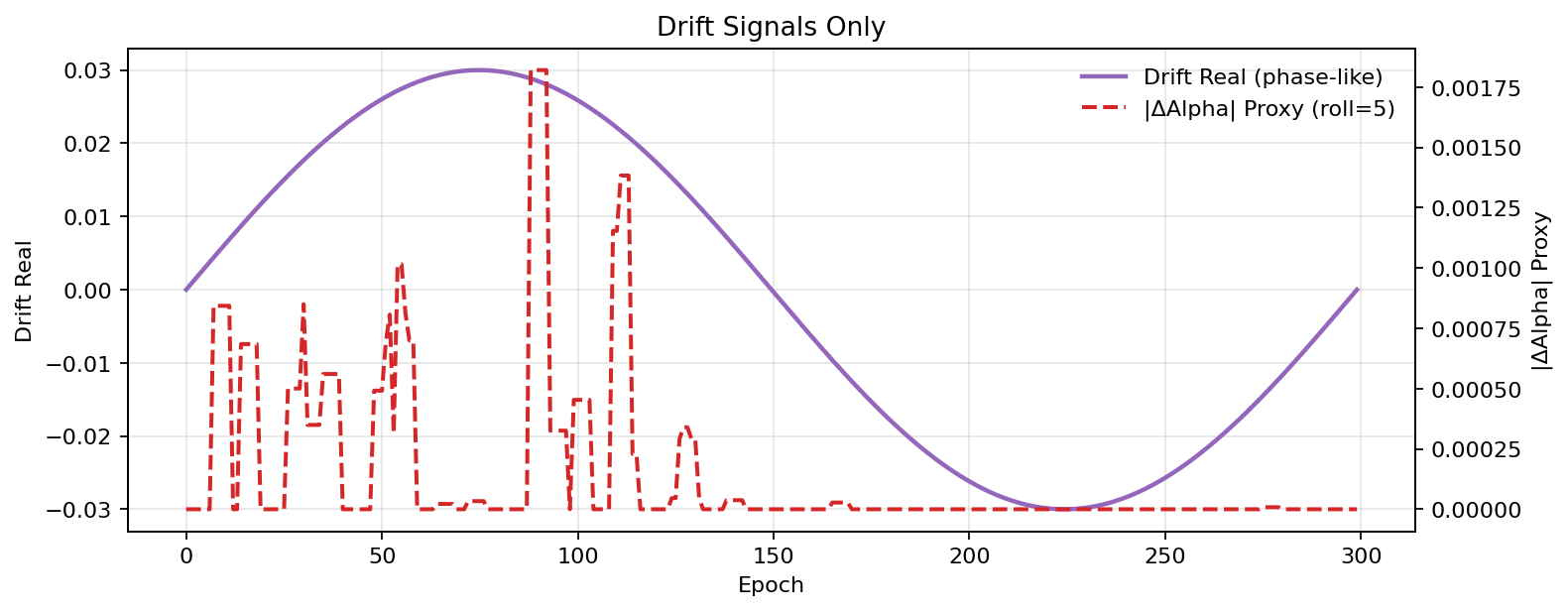}
  \caption{Real phase drift $\phi_t$ and smoothed proxy $|\Delta\alpha_t|$.}
  \label{fig:drift-signals}
\end{figure}

Peaks in the drift proxy coincide with regions where the sinusoidal drift
magnitude is highest.


\begin{figure}[H]
  \centering
  \includegraphics[width=0.48\textwidth]{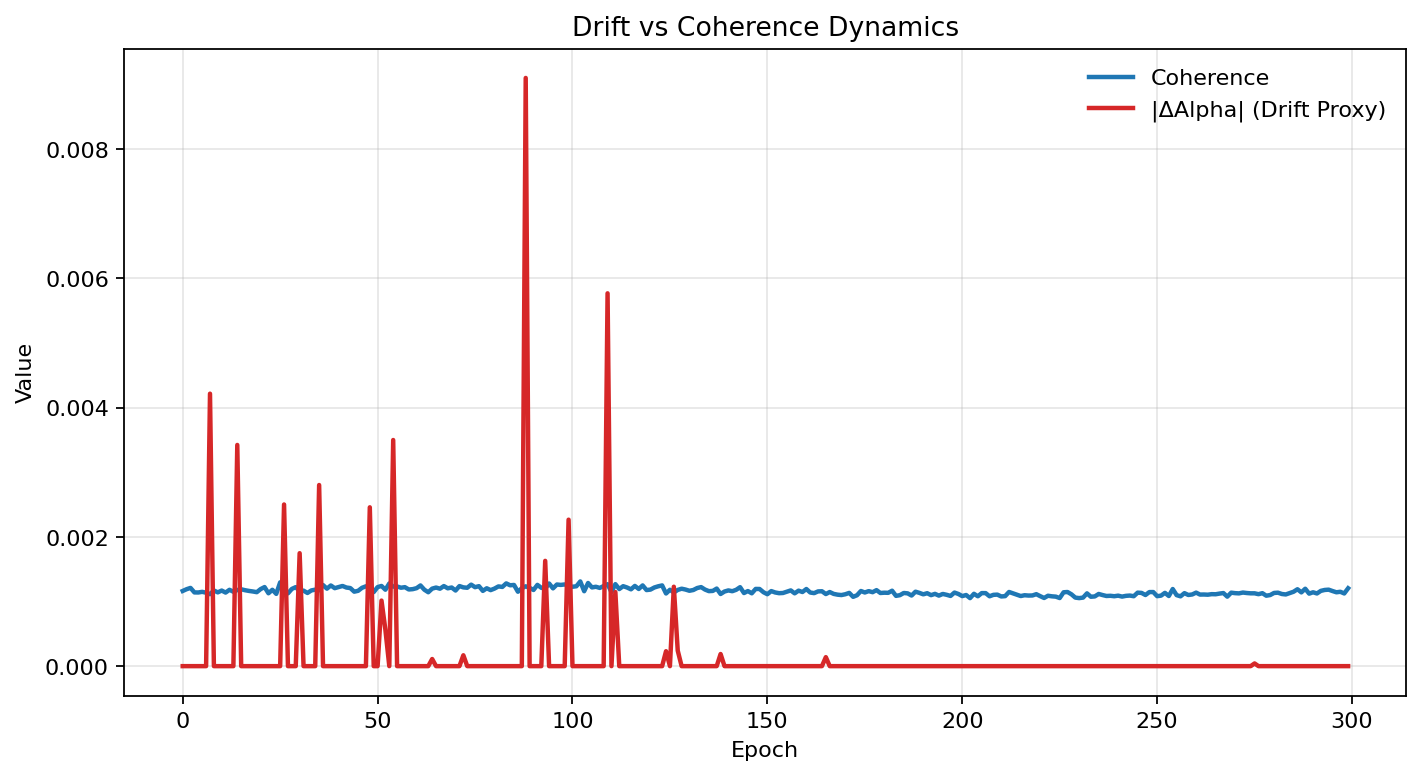}
  \caption{Coupled fluctuations between coherence and drift proxy spikes.}
  \label{fig:drift-vs-coherence}
\end{figure}

Periods where $|\Delta\alpha_t|$ spikes correspond to local perturbations in
coherence.

\subsection{Summary of Validation}

Collectively, these experiments validate that QML-HCS maintains stable
hypercausal behavior under structured physical drift. Losses remain bounded,
coherence and consistency evolve along a low-dimensional manifold, and the
feedback parameter tracks drift while converging to a stationary regime.

\section{Conclusion and Outlook}

QML-HCS introduces a unified hypercausal architecture that integrates quantum-inspired superposition principles, multi-branch causal propagation, and deterministic–stochastic hybrid execution into a single computational framework. The system demonstrates stable behavior under structured physical drift, maintains coherence and consistency across epochs, and exhibits robust adaptation through feedback-driven control parameters. Experimental results confirm that the hypercausal projection policy consistently preserves alignment between present states and representative futures, even under phase drift, detuning, and oscillatory readout perturbations.

The present release establishes the core infrastructure required for hypercausal computation: a typed execution model, a multi-backend processing layer, future-projection operators, metric-driven evaluation, and flexible optimization routines. Built in Python \cite{vanrossum2011python} and backed by efficient numerical primitives from NumPy \cite{Harris2020}, the framework is designed to be reproducible, accessible, and interoperable within modern scientific computing ecosystems. In this sense, QML-HCS already serves as \textit{a research-grade library for quantum-inspired machine learning with hypercausal feedback}.

Looking forward, several research directions emerge:

\textbf{Scalable Hypercausal Graphs.} Extending the system toward deep, recurrent, and hierarchical hypercausal architectures capable of long-horizon reasoning and multi-scale dynamical modeling.

\textbf{Quantum–Hybrid Benchmarks.} Systematic comparison with accelerator-backed differentiable frameworks such as JAX \cite{jax2018github}, alongside broader quantum-learning methodologies discussed in the literature on quantum machine learning \cite{schuld2021mlqc} and variational optimization techniques \cite{Preskill2018NISQ}.

\textbf{Mathematical Foundations.} Developing a formal theory of hypercausal dynamics, including stability criteria, causal manifolds, and drift-bounded propagation guarantees.

\textbf{Real-World Non-Stationary Applications.} Deploying QML-HCS in sensing, adaptive control, forecasting, edge-AI devices, and hybrid quantum–classical workflows that require continual adaptation to drift.

By providing a principled and extensible hypercausal learning framework, QML-HCS aims to lay a foundation for next-generation adaptive systems capable of operating coherently in dynamic, drift-dominated environments where traditional models struggle. Future releases will incorporate large-scale benchmarks, expanded backend support, and further theoretical refinement, strengthening QML-HCS as a comprehensive platform for quantum-inspired causal computation, quantum computing workflows and programming models targeting near-term quantum hardware, and autonomous machine-learning systems capable of self-correction and continual adaptation.

\section*{Acknowledgment}

The author gratefully acknowledges the NeureonMindFlux Research Lab for its
continuous support and for providing the computational resources used in the
design, implementation, and validation of the QML-HCS framework. Appreciation
is also extended to the broader open-source quantum and machine-learning
community for their foundational tools and resources.

\bibliographystyle{IEEEtran}
\bibliography{references}

\appendices

\section*{Appendix: Software and Resources}
{%
\hypersetup{
  urlcolor=blue,
  linkcolor=blue,
  citecolor=blue
}

To ensure full reproducibility and continued community development, all
artifacts associated with QML-HCS are publicly available:

\begin{itemize}
    \item \textbf{Source Code (GitHub):}  
    The complete implementation of the QML-HCS framework, including the core
    modules, backend adapters, hypercausal nodes, metrics, optimizers, and
    experiment scripts, is hosted at:  
    \url{https://github.com/Neureonmindflux-Research-Lab/qml-hcs}

    \item \textbf{PyPI Package (v0.2.0):}  
    The latest stable release of QML-HCS is available on the Python Package Index:  
    \url{https://pypi.org/project/qml-hcs/}  
    It can be installed via:  
    \texttt{pip install qml-hcs}

    \item \textbf{Archived Research Release (Zenodo):}  
    A versioned and citable research snapshot of the framework, including all
    source code and auxiliary materials, is available through Zenodo:  
    \url{https://doi.org/10.5281/zenodo.17562336}

    \item \textbf{Documentation (ReadTheDocs):}  
    Full API reference, architecture overview, tutorials, and examples can be
    found in the online documentation portal:  
    \url{https://qml-hcs.readthedocs.io/en/latest/}
\end{itemize}

These resources collectively support reproducible experimentation,
transparent validation, and continued extension of hypercausal learning
methods. Researchers are encouraged to contribute via pull requests, issue
reports, and extended applications of QML-HCS in quantum-inspired and
non-stationary machine-learning settings.
}

\end{document}